\documentclass[lettersize,journal]{IEEEtran}
\usepackage{amsmath,amsfonts}

\usepackage{array}
\usepackage[caption=false,font=normalsize,labelfont=sf,textfont=sf]{subfig}
\usepackage{textcomp}
\usepackage{stfloats}
\usepackage{url}
\usepackage{verbatim}
\usepackage{graphicx}

\usepackage{amsmath,amsfonts}
\usepackage{algorithm}
\usepackage{array}
\usepackage[caption=false,font=normalsize,labelfont=sf,textfont=sf]{subfig}
\usepackage{textcomp}
\usepackage{booktabs}
\usepackage{stfloats}
\usepackage{url}
\usepackage{verbatim}
\usepackage{orcidlink}
\usepackage{cite}
\usepackage{multirow}
\usepackage{cleveref}
\usepackage[numbers]{natbib}
\usepackage{textcomp}

\usepackage{algpseudocode}
\usepackage[numbers]{natbib}
\usepackage{textcomp}
\usepackage{hyperref} 
\usepackage{algpseudocode}
\usepackage{cite}
\usepackage{amsmath,amsfonts}
\usepackage{algorithm}
\usepackage{array}
\usepackage[caption=false,font=normalsize,labelfont=sf,textfont=sf]{subfig}
\usepackage{textcomp}
\usepackage{booktabs}
\usepackage{stfloats}
\usepackage{url}
\usepackage{verbatim}
\usepackage{orcidlink}
\usepackage{cite}
\usepackage{multirow}
\usepackage{cleveref}

\def\BibTeX{{\rm B\kern-.05em{\sc i\kern-.025em b}\kern-.08em
    T\kern-.1667em\lower.7ex\hbox{E}\kern-.125emX}}
\usepackage{balance}

\hypersetup{hidelinks}

\begin{document}
\title{Local Descriptors Weighted Adaptive Threshold Filtering For Few-Shot Learning}

\author{ \href{https://orcid.org/0000-0000-0000-0000}{\hspace{1mm}Bingchen Yan}\thanks{} \\
	Department of Computer Science\\
	Guilin University Of Electronic Technology\\
	Guilin, Gungxi, China  \\
	\texttt{1321847667a@gmail.com} 


}


\maketitle

\begin{abstract}
Few-shot image classification is a challenging task in the field of machine learning, involving the identification of new categories using a limited number of labeled samples. In recent years, methods based on local descriptors have made significant progress in this area. However, the key to improving classification accuracy lies in effectively filtering background noise and accurately selecting critical local descriptors highly relevant to image category information.

To address this challenge, we propose an innovative weighted adaptive threshold filtering (WATF) strategy for local descriptors. This strategy can dynamically adjust based on the current task and image context, thereby selecting local descriptors most relevant to the image category. This enables the model to better focus on category-related information while effectively mitigating interference from irrelevant background regions.

To evaluate the effectiveness of our method, we adopted the N-way K-shot experimental framework. Experimental results show that our method not only improves the clustering effect of selected local descriptors but also significantly enhances the discriminative ability between image categories. Notably, our method maintains a simple and lightweight design philosophy without introducing additional learnable parameters. This feature ensures consistency in filtering capability during both training and testing phases, further enhancing the reliability and practicality of the method.

\end{abstract}


\section{Introduction}
\IEEEPARstart{D}{eep} learning Deep learning models have achieved remarkable success across various computer vision domains when trained on large-scale manually annotated datasets [1]–[5]. However, these models continue to face significant challenges when dealing with novel classes containing only a few labeled samples, often resulting in overfitting or convergence failure. In contrast, humans can effortlessly recognize new classes from a limited number of labeled samples by leveraging prior knowledge. Few-shot learning aims to bridge this gap by generalizing knowledge acquired from base classes (with abundant labeled samples) to novel classes (with limited labeled samples), thus garnering increasing attention \cite{song2023learning,li2020more,zheng2023bdla,li2019revisiting,sun2024klsanet,zhou2024global,hao2021global,finn2017model,lee2019meta,wang2024fast,snell2017prototypical,vinyals2016matching,huang2021local,qi2022task,sung2018learning}.

The field has witnessed the emergence of various exemplary few-shot learning methods, broadly categorized into three types: metric learning-based \cite{li2019revisiting,zheng2023bdla,sun2024klsanet,snell2017prototypical,vinyals2016matching,huang2021local,li2020more,qi2022task,sung2018learning}, meta-learning-based \cite{finn2017model,lee2019meta,wang2024fast}, and transfer-based \cite{chen2024featwalk,fu2021meta,hu2022adversarial,gao2024few,tseng2020cross,chen2021meta,sun2021explanation} approaches. Notably, metric learning-based methods have achieved significant success due to their simplicity and efficacy. This paper primarily focuses on this approach. The typical pipeline of metric learning-based few-shot learning methods encompasses three steps: 1) Feature extraction from all query and support images; 2) Distance computation between the query image and each support image, prototype, or class center using a specific metric; 3) Label assignment to query images through nearest neighbor search.
\begin{figure}[ht]
\centering
\includegraphics[scale=0.45]{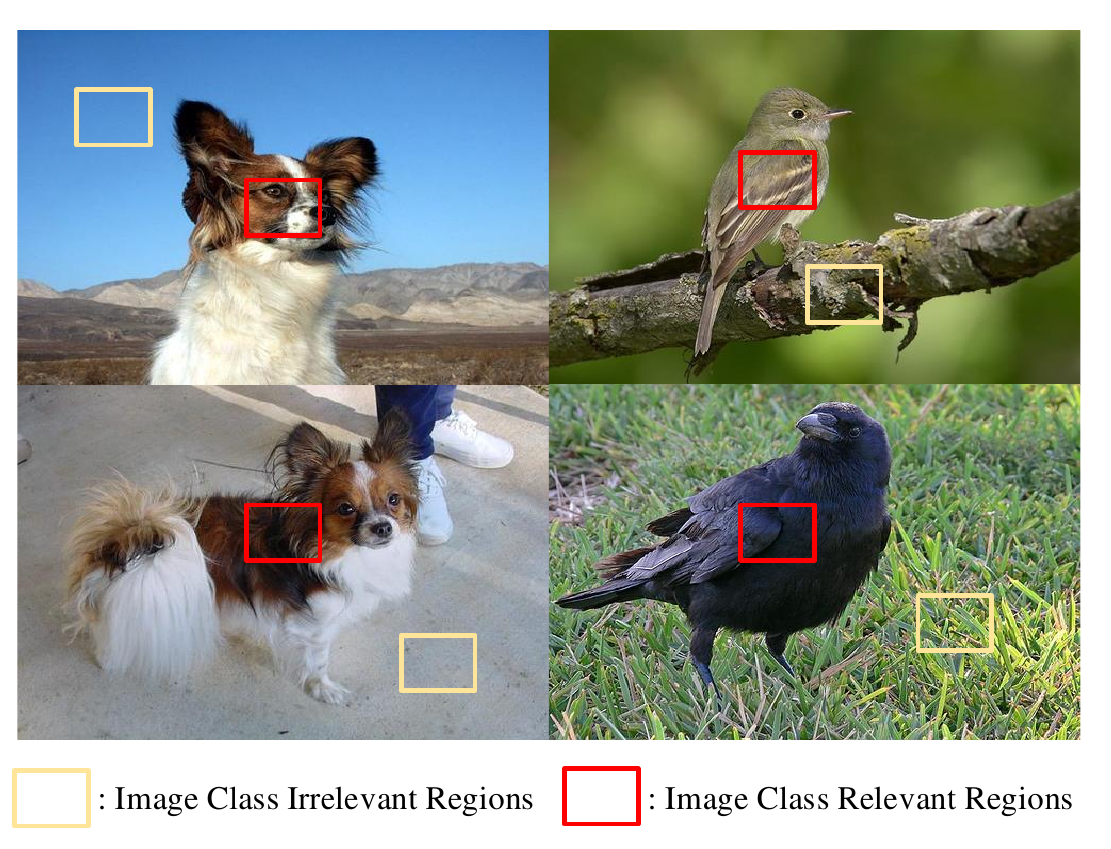}
\caption{Examples of regions that are relevant and irrelevant to image classes.}
\label{region}
\end{figure}
Despite the impressive performance of metric learning-based few-shot learning methods, they are persistently plagued by noisy local regions irrelevant to image category information, as the semantics of local regions within images can vary significantly\cite{dong2021learning,qi2022task}. As illustrated in Figure \ref{region}, some regions contain critical semantics consistent with image category information, i.e., category-relevant information (e.g., the "dog" area in a "dog" image, or the "bird" area in a "bird" image). Conversely, other regions may contain semantics irrelevant to image category information, i.e., category-irrelevant information (e.g., the "sky" area in a "dog" image, or the "grass" area in a "bird" image).

To address this issue, GLIML\cite{hao2021global} and KLSANet \cite{sun2024klsanet}employ a dual-branch architecture to simultaneously learn global and local features of images, selecting local features by measuring their similarity to the image's global features. Although this approach has yielded significant results, it substantially increases model complexity and computational time. BDLA \cite{zheng2023bdla}proposed computing bidirectional distances between local features of query and support samples to enhance the effective alignment of contextual semantic information in image local features.

Our method builds upon previous work \cite{li2019revisiting,li2020more,zheng2023bdla,song2023learning} utilizing local descriptor-level image features. Aiming to maximize the elimination of noisy local regions irrelevant to image category information, we ingeniously propose a weighted dynamic filtering method for local descriptors in few-shot learning.

Specifically, we innovatively introduce the concept of category-relevant weights for local descriptors. Through visualization experiments, we demonstrate that these weights conform to a normal distribution in statistical terms. Based on this observation, we design an adaptive threshold strategy for weights, dynamically filtering out local regions with the highest relevance to category information.

In summary, our work makes the following key contributions:
\begin{itemize}

    \item 
We propose a novel weighted dynamic filtering method for local descriptors in few-shot learning, which effectively addresses the challenge of noisy, category-irrelevant regions in images.

    \item 
We introduce the concept of category-relevant weights for local descriptors and empirically demonstrate their conformity to a normal distribution through visualization experiments.

    \item 
Based on this statistical insight, we develop an adaptive threshold strategy that dynamically selects the most category-relevant local regions, significantly enhancing the model's focus on pertinent information.

    \item 
Our method achieves state-of-the-art performance on three widely-used few-shot learning classification datasets, surpassing existing metric learning-based approaches.
\end{itemize}
Our method outperforms current state-of-the-art approaches on three commonly used few-shot learning classification datasets. More surprisingly, the experimental results on the CUB-200 dataset even surpass several recent transfer learning-based few-shot learning methods. We believe our method holds significant reference value for subsequent research in few-shot learning.

\section{Related works}
Few-shot learning algorithms can be broadly categorized into three main classes: initialization-based methods, methods rooted in transfer learning and metric-based methods.

\subsection{Initialization-based methods}
Initialization-based methods \cite{finn2017model,lee2019meta,wang2024fast,rusu2018meta,triantafillou2019meta,wang2022global} utilize gradient updates to achieve effective initialization. MAML \cite{finn2017model} introduced a powerful initialization technique that significantly enhances performance with just a few gradient steps, employing a bi-level optimization strategy where the outer loop learns to generalize across tasks and the inner loop adapts to specific tasks. LEO \cite{rusu2018meta} extends MAML by operating in a low-dimensional space to improve generalization in FSL tasks. Proto-MAML \cite{triantafillou2019meta} combines the strong inductive bias of ProtoNet \cite{snell2017prototypical} with the flexible adaptation mechanism of MAML \cite{finn2017model}. However, the MAML family typically uses a simple cross-entropy function for inner loop optimization, which can result in limited generalization performance. To address this, Baik et al. (2021) \cite{baik2021meta} proposed a task-specific loss function to update meta-learner parameters during the meta-training process. Wang et al. (2022) \cite{wang2022global}provided a theoretical analysis of how MAML with deep neural networks converges to the global optimum and developed a specialized neural architecture search algorithm for FSL.

\subsection{Methods rooted in transfer learning}
Methods rooted in transfer learning frameworks have demonstrated competitive performance in the realm of few-shot learning, often rivaling meta-learning techniques. The general methodology of these approaches follows a distinct pattern:

Initially, a classification model is trained on the entire available training dataset.
Subsequently, the classification layer is discarded, preserving only the feature extraction component.
Finally, utilizing the support set from the test data, a new classifier is developed and trained.
This strategy has proven effective, with several notable implementations gaining traction in the field. Among these, Dynamic Classifier [28], Baseline++ \cite{fu2021meta}, and RFS [30] stand out as particularly influential contributions.

\subsection{Metric-Based Methods}

Metric-based methods \cite{li2019revisiting,zheng2023bdla,sun2024klsanet,snell2017prototypical,vinyals2016matching,huang2021local,li2020more,qi2022task,sung2018learning} aim to learn a universal metric space to measure the relationship between query images and support sets, thereby quantifying their similarity. Matching Networks \cite{vinyals2016matching} determine the similarity between each support set sample and a query sample, predicting the query sample's label by computing a weighted sum of these similarities. Prototypical Networks innovatively average the support set features to form class prototypes and evaluate the Euclidean distance between the query and class prototypes in the embedding space \cite{snell2017prototypical}. Relation Networks compare the relation between images by learning a deep nonlinear metric. TADAM \cite{oreshkin2018tadam}enhances few-shot learning (FSL) by learning a task-dependent metric space through metric scaling.

Despite their potential, current methods largely depend on image-level global features, assuming their transferability across seen and unseen classes, which is often unrealistic. In contrast, low-level features like local descriptors and local features are more likely to be shared among different classes and are expected to transfer better to unseen classes  have demonstrated the superiority of local descriptors over global representations in few-shot image classification.

For instance, LMP-Net \cite{huang2021local}leverages local descriptor-level features rather than global features in Prototypical Networks, learning multiple class prototypes for each class to capture the complex distribution of the class more comprehensively. DN4 \cite{li2019revisiting} employs deep local descriptor representation and explicitly uses local descriptors through $k$-nearest neighbors ($k$-NN), while the Relational Network \cite{sung2018learning} implicitly measures distances between query and support samples using local descriptors. However, local descriptors often contain redundant information from spatially adjacent areas, and the semantic local descriptors commonly shared by all classes are not crucial for recognizing new instances \cite{dong2021learning}.
To address the limitations of local descriptor-based methods, ATL-Net \cite{dong2021learning}designs an episodic attention mechanism that can select and weight key local descriptors without overemphasizing the common parts across the entire task. BDLA \cite{zheng2023bdla} introduces the calculation of bidirectional distance between local descriptors of query samples and support samples to enhance the effective alignment of contextual semantic information. KLSANet \cite{sun2024klsanet}utilizes randomly cropped local features instead of local descriptors, selecting key query local features by measuring their relationship to the image semantics to reduce the impact of irrelevant query parts on image semantics. However, extracting both local features and global feature representations for each image significantly increases computational overhead and model complexity.

\section{mehtod}

\subsection{Problem Definition}

Few-shot learning aims to develop models that excel with minimal data while maintaining robust generalization. We tackle the N-way K-shot challenge, where N represents class count and K denotes samples per class, typically a small number like 1 or 5.

Our goal is to train model parameters $\theta$, for swift adaptation to unseen data using episodic training. Each episode in both training and test datasets contains a support set S (N classes, K labeled images each) and a query set Q for evaluation.

The data is split into non-overlapping training, validation, and testing sets, each containing more classes and samples than N and K. These sets are then further divided into episodes with distinct support and query sets sharing the same label space.

To simulate real-world scenarios, all phases employ this episodic mechanism. For example, during training, random episodes are selected for parameter updates until convergence. In validation and testing, the model classifies the query set based on the support set.

\subsection{Overview}

As illustrated in Figure [X], our proposed approach comprises three principal components: the Embedding Feature Extraction Module (EFEM), the Weighted Adaptive Threshold Filtering Module (WATFM), and the 
 Key Local Descriptors Classification Module (KLDCM).

Initially, we employ an embedding network constructed on the episodic learning mechanism to extract local descriptor-level embedding features from both the support set and query set images. Subsequently, the WATFM computes weight information for each local descriptor of the images in the support and query sets. This process enables the identification and selection of key local descriptors while eliminating background noise, thereby enhancing few-shot classification performance.

In the final stage, we input the filtered key local descriptors from both the support and query set images into a $k$-Nearest Neighbors ($k$-NN) classifier, a commonly used technique in previous works. This classifier then generates the predicted class labels for the query set images.

\subsection{EFEM}

We utilize a widely-used neural network, typically a Convolutional Neural Network (CNN) or ResNet, following previous work, to serve as a local descriptor feature extractor. This local descriptor feature extractor can be implemented by removing the last pooling layer or the fully connected layer of the neural network. To illustrate with a CNN as an example:

Each image \( X \) is passed through the CNN to obtain a three-dimensional (3D) tensor \( \mathcal{F}_\theta(X) \in \mathbf{R}^{C \times H \times W} \). This tensor represents the image, where \( \mathcal{F}_\theta(X) \) is the hypothesized function learned by the CNN, \( \theta \) stands for the parameters of the CNN, and \( C \), \( H \), and \( W \) denote the channel, height, and width of the 3D tensor, respectively. This can be expressed as:

\begin{equation}
    \mathcal{F}_\theta(X) = \left[\boldsymbol{x}^1, \ldots, \boldsymbol{x}^M\right] \in \mathbf{R}^{C \times M}
\end{equation}

Here, \( M = H \times W \), maps all images to a representational space. Each 3D tensor contains \( M \) units of \( C \) dimensions, with each unit representing a local descriptor of the image.

\subsection{WATFM}
Due to the large intra-class variation and background clutter, the measurement of using all local descriptors directly for few-shot image classification is far from satisfactory. Therefore, it is more reasonable to filter out the local descriptors most relevant to the category and then carry out subsequent operations.

Our local descriptor filtering strategy is based on the following premise: As shown in Figure X, in a typical few-shot task, the support set usually consists of five categories, with N typically set to 5. For $K$ support set images of a category, if a local descriptor in one of the $K$ support set images is category-relevant (containing exact representative features of that category), then similar local descriptors should exist in the other $(K-1)$ support set images. Conversely, if a local descriptor comes from a background area irrelevant to the category of the support set image, the likelihood of similar local descriptors appearing in the other $(K-1)$ support set images of the same category is low, and they may even appear in support set images of other categories.

Following the approach of ProtoNet \cite{snell2017prototypical}, we calculate the category prototype for each support set category by averaging, which possesses more comprehensive and representative information related to the support set category, used for key local descriptor filtering. The filtering process includes two main steps. First, we compute the similarity between each candidate local descriptor of the support sample and its support set category prototype. In the feature embedding space, we denote the prototype representation of the nth category as $C_n$, where n $\in$ [1, N].

For a support set image, we obtain the local descriptor representation as follows:

\begin{equation}
    \mathcal{F}_\theta(X_S) = \left[\boldsymbol{x}^1_S, \ldots, \boldsymbol{x}^M_S\right] \in \mathbf{R}^{C \times M}
\end{equation}

where ${x}^i_S$, i $\in$ [1, M] represents the ith local descriptor extracted by EFEM belonging to the support set image, and M represents the number of local descriptors.

We then calculate the similarity between each category prototype and each local descriptor using the following formula:

The importance $\omega_{i, n}$ of the local descriptor $\boldsymbol{x}^i_S$ for the $n$-th class can be estimated by the normalized cosine similarity between the local descriptor $\boldsymbol{x}^i_S$ and the prototype representation $\mathbf{c}_n$, i.e.,

\begin{equation}
\omega_{i, n}=\frac{e^{ \cos\left(\boldsymbol{x}^i_S, \mathbf{c}_n\right)}}{\sum_{i=1}^M e^{ \cos\left(\boldsymbol{x}^i_S, \mathbf{c}_n\right)}},
\end{equation}

As seen from Equation 3, local descriptors containing category-relevant information for the kth class will have higher importance weights, while those containing category-irrelevant information will have lower weights.

Based on the calculated weights, we select local descriptors with high weights for subsequent processing while ignoring those with low weights. We accomplish this by setting an adaptive threshold that automatically adjusts according to the weight distribution, retaining only local descriptors with weights above this threshold. Our threshold filtering strategy adapts to the number of key local descriptors, dynamically changing according to the current task and different image local contexts.

Specifically, through Equation (3), we calculate a weight matrix $\mathbf{W}$ with shape $[L, N, M]$, where $L$ represents the number of support set or query set samples, $N$ represents the number of categories, and $M$ represents the number of local features per sample.

\textbf{Weight Aggregation and Expansion:}
In our method, the weight of each local descriptor is five weights for five categories, each weight corresponding to a specific category. To derive the importance of each local descriptor across all five categories, we need to average the weight values of the five categories, thus obtaining the importance of this local descriptor for the five categories, i.e., whether this local descriptor is important for the main subjects of images across all five categories.

The formula is as follows:

\begin{equation}
\overline{w}_{i} = \frac{1}{N} \sum_{n=1}^{N} w_{i,n}
\end{equation}

where $\overline{w}_{i}$ represents the average weight of the ith local descriptor, $N$ represents the number of categories, $w_{i,n}$ represents the weight of the ith local descriptor for the nth category, where n $\in$ [1, N], i $\in$ [1, M].

Threshold Calculation:
To determine the adaptive threshold, we first calculate the mean and standard deviation of the average weights of all local descriptors:

\begin{equation}
\mu = \frac{1}{L \times M} \sum_{j=1}^{L} \sum_{i=1}^{M} \overline{w}_{i,j}
\end{equation}
\begin{equation}
\sigma = \sqrt{\frac{1}{L \times M} \sum_{j=1}^{L} \sum_{i=1}^{M} (\overline{w}_{i,j} - \mu)^2}
\end{equation}

where $\mu$ represents the mean of the average weights of all local descriptors, $\sigma$ represents the standard deviation of the average weights of all local descriptors, and $\mathbf{w}_{i,j}$ represents the average weight of the ith local descriptor of the jth support or query sample.
\begin{figure}[ht]
\centering
\includegraphics[scale=0.6]{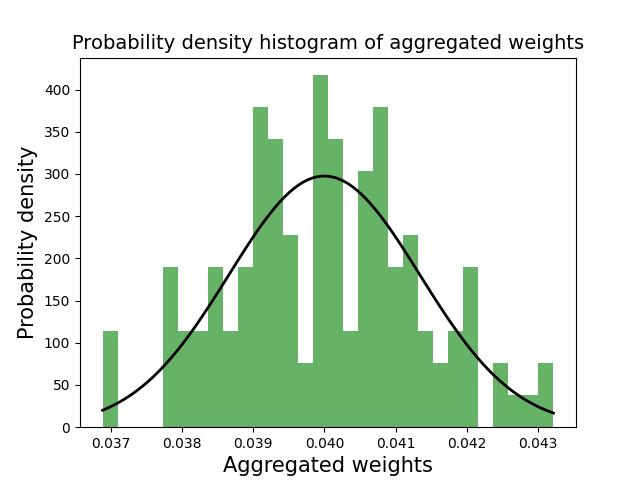}
\caption{Visualization of probability distribution histogram of the average weights of local descriptors for support set images.}
\label{zhengtai}
\end{figure}
In statistics, the 68-95-99.7 rule is an empirical rule in normal distribution, indicating that in a normal distribution, the proportion of data within one, two, and three standard deviations from the mean is 68.27\%, 95.45\%, and 99.73\%, respectively. Taking the experiment conducted on the Stanford Dogs dataset under the 1-shot experimental setting as an example, we plotted a probability distribution histogram of the average weights of local descriptors for support set images, as shown in Figure \ref{zhengtai}. We found that it follows a normal distribution, thus in our study, we utilized this rule to determine the filtering threshold for local descriptors.

Specifically, we define the adaptive threshold $\tau$ as the mean minus one standard deviation:

\begin{equation}
\tau = \mu - \sigma
\end{equation}
This corresponds to the part in the normal distribution that is greater than one standard deviation from the mean, which accounts for approximately 15.87\% ((100\% - 68.27\%) / 2) of the total. Therefore, we actually retain about 84.14\% of the local descriptors, whose weights are higher than or equal to the mean and can be considered more important parts for the main subjects of images across the five categories.

Filtering Strategy:
Based on the calculated threshold $\tau$, we retain all local descriptors with weights higher than $\tau$:

\begin{equation}
\mathcal{F}_{\theta_{\text{filtered}}}(\overline{X}) = \{\overline{\boldsymbol{x}}^i_j \mid \overline{w}_{i,j} > \tau \}
\end{equation}

where $\mathcal{F}_{\theta_{\text{filtered}}}(\overline{X})$ represents the set of filtered local descriptors, $\overline{\boldsymbol{x}}^i_j$ represents the ith filtered local descriptor of the jth support or query sample.
\begin{figure*}[ht]
\centering
\includegraphics[scale=0.6]{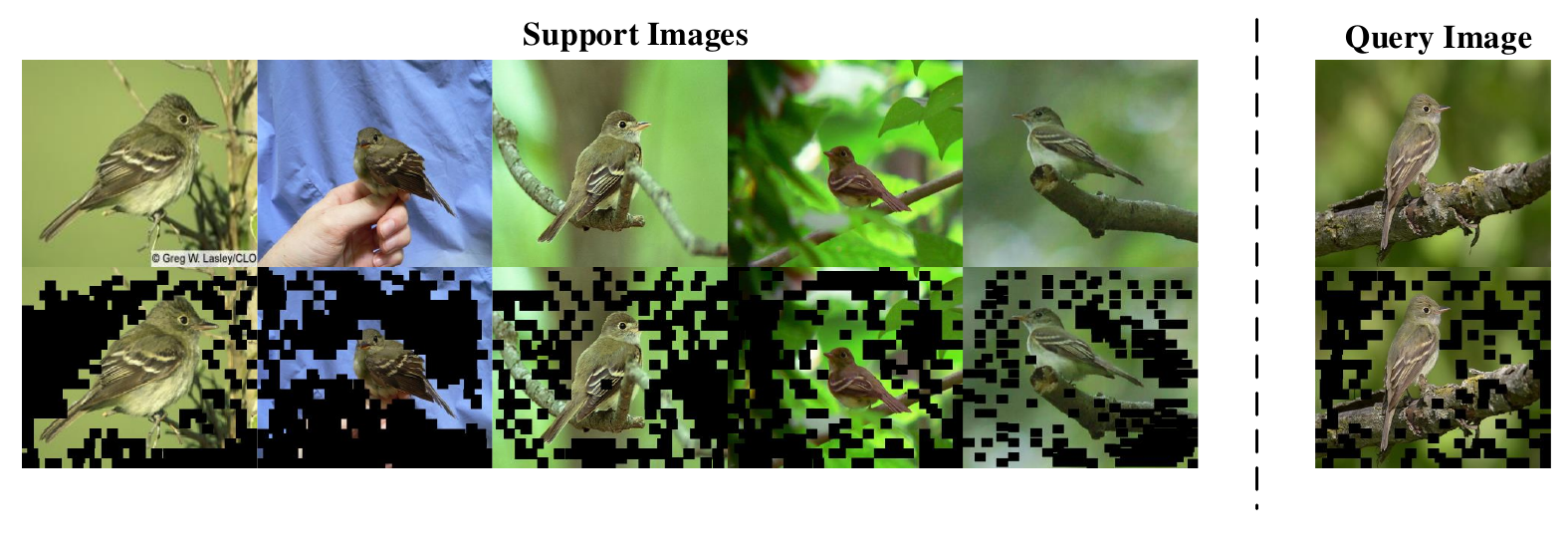}
\caption{Visualization results of local descriptors for four randomly sampled 5-way 1-shot classification tasks, comparing the cases with and without our WATF module.}
\label{masaike}
\end{figure*}
Figure \ref{masaike} shows the visualization results of local descriptors for four randomly sampled 5-way 1-shot classification tasks, comparing the cases with and without our WATF module. As shown in the Figure \ref{julei}, when using WATF, the selected local descriptors within each cluster exhibit a more compact arrangement, indicating that it is easier to distinguish local descriptors between different image categories.

After filtering the local descriptors of the support set through the above steps, we recalculate the category prototypes and repeat the above local descriptor filtering on the query set using the updated category prototypes. The algorithm flow is shown in Algorithm Pseudocode X.

Through our WATF module, the neural network can focus attention on the category-relevant key information of the image, improving the representation of support set and query set images, mitigating the negative impact of category-irrelevant non-target areas. Moreover, our filtering method maintains simplicity and lightweight design without introducing additional learnable parameters, ensuring consistency in filtering capability during both training and testing phases.

\subsection{KLDCM}

To predict the category of a query image, we extend the concept of image-to-class measure, utilizing the selected local descriptors for classification. Specifically,

The key local descriptors of a given query image q selected after WATFM filtering are represented as:

\begin{equation}
    \mathcal{F}_{\theta_{\text{filtered}}}(\overline{X_q}) = \left[\overline{\boldsymbol{x}}^1_q, \ldots, \overline{\boldsymbol{x}}^H_q\right] \in \mathbf{R}^{C \times H}
\end{equation}

where \( H \leq M \). After WATFM filtering, each category in the support set can be represented as class \( i \) ($i=1, 2, 3, \cdots, 5$). For each filtered key local descriptor $\overline{\boldsymbol{x}}^h_q$ of q, where h $\in$ [1,H], we find its \( k \) nearest neighbors denoted as \( n_1, \cdots, n_k \) in each filtered support set local descriptor and compute the corresponding cosine similarities as \( \cos(\overline{\boldsymbol{x}}^h_q, n_1), \cdots, \cos(\overline{\boldsymbol{x}}^h_q, n_k) \). The similarity score between image \(q \) and class \( i \) is defined as:

\begin{equation}
\begin{aligned}
    \operatorname{Score}(q, \operatorname{class} i) = \sum_{h=1}^H \sum_{j=1}^k \cos(\overline{\boldsymbol{x}}^h_q, n_j)
\end{aligned}
\end{equation}

Then, we use softmax to obtain the probability that the category $y_q$ of $\mathbf{q}$ is class $i$:

$$
p(y_q=i \mid \mathbf{q}) = \frac{\exp \left(\operatorname{score}\left(q, \operatorname{class} i\right)\right)}{\sum_{i=1}^5 \exp \left(\operatorname{score}\left(q, \operatorname{class} i\right)\right)} .
$$

\section{experiment}

\subsection{Datasets}

\textbf{CUB-200} is a fine-grained bird image classification dataset involving 200 different bird species. The number of images per category varies, with 130 categories used for training, 20 for validation, and the remaining 50 for testing.

\textbf{The Stanford Dogs} dataset focuses on fine-grained dog image classification, comprising 20,580 photographs of 120 different dog breeds. 70 dog breeds are used for training, 20 for validation, and the remaining 30 for testing.

\textbf{The Stanford Cars} dataset is designed for fine-grained car image classification, containing 16,185 images of 196 different car categories, defined by make, model, and year of manufacture. 130 categories are used for training, 17 for validation, and the remaining 49 for testing.

\subsection{Implementation Details
}
In our experiments, we primarily focus on 5-way 1-shot and 5-shot classification tasks. To ensure fair comparison with other methods, we employ two commonly used backbone network structures in few-shot learning: Conv4 and ResNet-12, following the implementation details outlined in DN4 \cite{li2019revisiting}and CovaMNet \cite{li2019distribution}.

During the training phase, we use the Adam optimization algorithm (Kingma \& Ba, 2014) with an initial learning rate of 0.001, which is halved every 100,000 episodes.

In the testing phase, to ensure the reliability of the experimental results, we randomly construct 600 episodes from the test set of each dataset to evaluate the model's performance. We select the best model based on the accuracy on the validation set and then evaluate it on the test set, which contains new classes. Each randomly sampled new task from the test set is similar to the training tasks, containing 5 classes, with K (1 or 5) support samples per class and 15 query samples per class. The test results are reported as the mean accuracy over 600 new tasks with a 95\% confidence interval. It is worth noting that our model is trained end-to-end from scratch, with no fine-tuning performed during the testing phase.

\subsection{Experimental Results
}

\subsubsection{General few-shot classification}

To validate the effectiveness and superiority of our proposed WATFM method, we compare our approach with 14 state-of-the-art few-shot classification methods on three fine-grained datasets, as summarized in Table \ref{tab1}.

It can be observed that WATF with ResNet-12 backbone significantly outperforms all comparison methods on most settings across the three datasets. Benefiting from less noisy local features, it can more accurately depict discriminative regions, showing significant improvements compared to other methods. In 1-shot and 5-shot settings, even using the same four-layer convolution as a local feature extractor, WATF improves accuracy by an average of 9.27\% and 2.75\% compared to the DN4 method that does not process local descriptors. This reveals to some extent how poor local descriptor representations can degrade classification performance in fine-grained image classification scenarios.

Notably, existing metric-based few-shot image classification methods can be divided into three categories based on feature level: global feature-based methods (e.g., ProtoNets \cite{snell2017prototypical}, GNN \cite{satorras2018few}, and QPN \cite{abdelaziz2023learn}), local descriptor-based methods (e.g., DN4 \cite{li2019revisiting}, DN4-DA \cite{li2019revisiting}, RelationNet \cite{sung2018learning}, MADN4 \cite{li2020more}, TDSNet \cite{qi2022task}, LMPNet \cite{huang2021local}, ATL-Net \cite{dong2021learning}, CovaMNet \cite{li2019distribution}, and BDLA \cite{zheng2023bdla}), local random crop feature-based methods (e.g., KLSANet \cite{sun2024klsanet}), and methods combining global features and local descriptors (e.g., GLCL\cite{zhou2024global}, GLIML \cite{hao2021global}). The recent excellent performance of local random crop feature-based methods once raised doubts about whether local descriptors were too detailed, losing crucial image local semantic information. Our method's outstanding performance reaffirms the superior position of local descriptor-level features in few-shot image classification.

Specifically, in 1-shot and 5-shot settings, WATF improves by an average of 10.05\% and 4.91\% compared to the best global representation-based method QPN, and by an average of 8.77\% and 4.71\% across the three datasets compared to the local feature-based method KLSANet. Comparing the experimental results with methods like BDLA and Hao, which also focus on improving local descriptor semantic alignment [BDLA, Hao], demonstrates the superiority of our method that does not introduce additional learnable parameters.
\begin{figure*}[ht]
\centering
\includegraphics[scale=0.6]{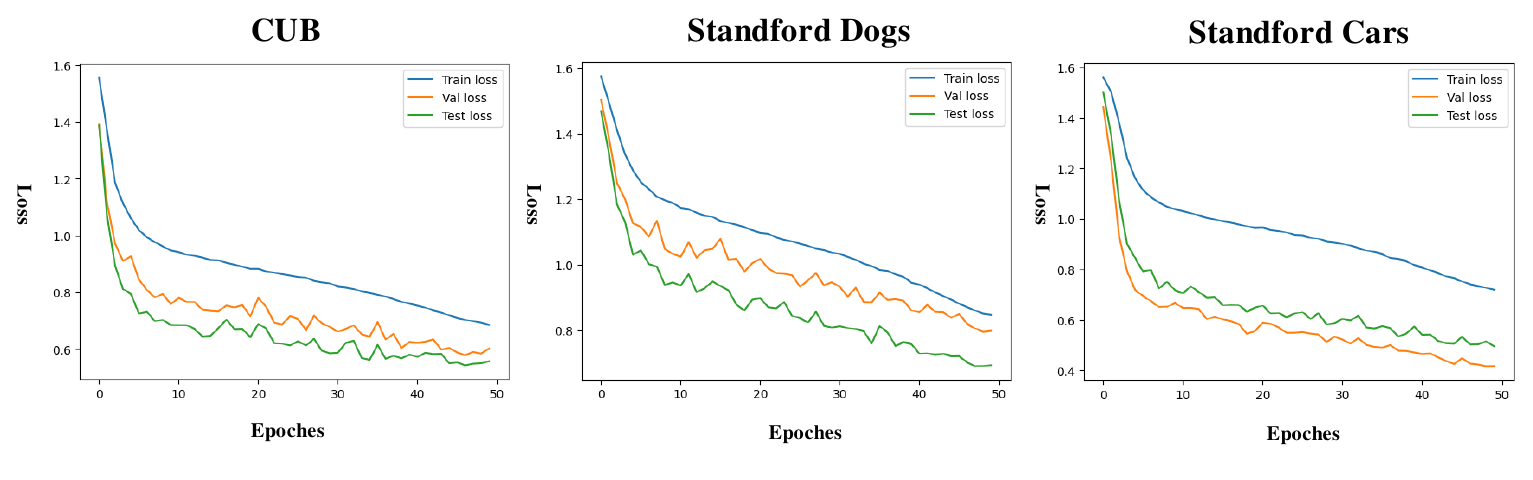}
\caption{Training loss, validation loss, and test loss curves of the proposed WATF on 5-way 1-shot setting of three datasets.}
\label{loss}
\end{figure*}
\textbf{Convergence Analysis.}To analyze the convergence of WATF, we present its training loss, validation loss, and test loss curves under the 5-way 1-shot setting across three datasets in Figure \ref{loss}. Across all three datasets, we observe that around the 50th epoch, the test loss stops decreasing, indicating model convergence. This demonstrates that our model is optimally trained and able to converge rapidly. Furthermore, we employ validation accuracy for model weight file selection in experiments to avoid overfitting.

\begin{table*}[!t]
\caption{Comparison with state-of-the-art methods on three fine-grained datasets, i.e., CUB, Stanford Dogs and Stanford Cars. Accuracies are reported with $95 \%$ confidence intervals. The results of the optimal and suboptimal comparison methods are bolded and underlined respectively.}
\label{tab1}
\centering
\begin{tabular}{cccccccc}
\toprule
\multirow{2}{*}{Method} & \multirow{2}{*}{Backbone} & \multicolumn{2}{c}{CUB} & \multicolumn{2}{c}{Stanford dogs} & \multicolumn{2}{c}{Stanford cars} \\
\cmidrule{3-4} \cmidrule{5-6} \cmidrule{7-8}
& & 1-shot & 5-shot & 1-shot & 5-shot & 1-shot & 5-shot \\
\midrule

 ProtoNets & Conv-4 & $51.31 \pm 0.91$ & $70.77 \pm 0.69$ & $37.80 \pm 0.99$ & $48.19 \pm 1.03$ & $40.90 \pm 1.01$ & $52.93 \pm 1.03$ \\
 RelationNet & Conv-4 & $62.45 \pm 0.98$ & $76.11 \pm 0.69$ & $43.33 \pm 0.42$ & $55.23 \pm 0.41$ & $47.67 \pm 0.47$ & $60.59 \pm 0.40$ \\
 GNN & Conv-4 & $51.83 \pm 0.98$ & $63.69 \pm 0.94$ & $46.98 \pm 0.98$ & $62.27 \pm 0.95$ & $55.85 \pm 0.97$ & $71.25 \pm 0.89$ \\
 QPN & Conv-4 & $66.04 \pm 0.82$ & $82.85 \pm 0.76$ & $53.69 \pm 0.62$ & $70.98 \pm 0.70$ & $63.91 \pm 0.58$ & $89.27 \pm 0.78$ \\
 DN4 & Conv-4 & $46.84 \pm 0.81$ & $74.92 \pm 0.64$ & $45.41 \pm 0.76$ & $63.51 \pm 0.62$ & $59.84 \pm 0.80$ & $88.65 \pm 0.44$ \\
 DN4-DA & Conv-4 & $53.15 \pm 0.84$ & $81.90 \pm 0.60$ & $45.73 \pm 0.76$ & $66.33 \pm 0.66$ & $61.51 \pm 0.85$ & $\textbf{89.60} \pm \textbf{0.44}$ \\
 CovaMNet & Conv-4 & $52.42 \pm 0.76$ & $63.76 \pm 0.64$ & $49.10 \pm 0.76$ & $63.04 \pm 0.65$ & $56.65 \pm 0.86$ & $71.33 \pm 0.62$ \\
 ATL-Net & Conv-4 & $60.91 \pm 0.91$ & $77.05 \pm 0.67$ & $54.49 \pm 0.92$ & $73.20 \pm 0.69$ & $67.95 \pm 0.84$ & $89.16 \pm 0.48$ \\
 MADN4 & Conv-4 & $57.11 \pm 0.70$ & $77.83 \pm 0.40$ & $50.42 \pm 0.27$ & $70.75 \pm 0.47$ & $62.89 \pm 0.50$ & $89.25 \pm 0.34$ \\
 TDSNet & Conv-4 & $\underline{69.34 \pm 0.89}$ & $80.34 \pm 0.59$ & $54.48 \pm 0.87$ & $69.45 \pm 0.69$ & $62.14 \pm 0.91$ & $75.64 \pm 0.72$ \\
 BDLA & Conv-4 & $50.59 \pm 0.97$ & $75.36 \pm 0.72$ & $48.53 \pm 0.87$ & $70.07 \pm 0.70$ & $64.41 \pm 0.84$ & $89.04 \pm 0.45$ \\
 AGLRs & Conv-4 & $\underline{69.34 \pm 0.70}$ & $\underline{84.72 \pm 0.42}$ & $58.85 \pm 0.69$ & $\underline{75.82 \pm 0.49}$ & $70.71 \pm 0.66$ & $\underline{89.42 \pm 0.33}$ \\
 KLSANet & Conv-4 & $66.70 \pm 0.82$ & $83.63 \pm 0.28$ & $52.23 \pm 0.56$ & $70.45 \pm 0.37$ & $54.71 \pm 0.77$ & $78.47 \pm 0.57$ \\
 ours & Conv-4 & $65.94 \pm 0.97$ & $79.96 \pm 0.52$ & $56.00 \pm 0.68$ & $73.70 \pm 0.54$ & $57.95 \pm 0.61$ & $81.66 \pm 0.45$ \\
 
 LMPNet & ResNet-12 & $65.59 \pm 0.68$ & $68.19 \pm 0.23$ & $\underline{61.89 \pm 0.10}$ & $68.21 \pm 0.11$ & $\underline{68.31 \pm 0.45}$ & $80.27 \pm 0.23$ \\
 KLSANet & ResNet-12 & $\textbf{74.94} \pm \textbf{0.43}$ & $\textbf{88.92} \pm \textbf{0.41}$ & $\textbf{64.43} \pm \textbf{0.81}$ & $\textbf{81.07} \pm \textbf{0.31}$ & $\textbf{74.43} \pm \textbf{0.76}$ & $87.84 \pm 0.45$ \\
 ours & ResNet-12 & $\textbf{79.63} \pm \textbf{0.64}$ & $\textbf{91.18} \pm \textbf{0.35}$ & $\textbf{74.80} \pm \textbf{0.69}$ & $\textbf{85.27} \pm \textbf{0.44}$ & $\textbf{85.41} \pm \textbf{0.60}$ & $\textbf{95.61} \pm \textbf{0.28}$ \\

\bottomrule
\end{tabular}
\end{table*}

\begin{table*}[!t]
   \caption{Cross-domain performance comparison of the proposed WATF with state-of-the-art methods on miniImageNett→CUB setting. ‘–’: not reported.}
\label{tab2}
\centering
\begin{tabular}{lllr}
 \hline Method & Backbone & miniImageNet $\rightarrow$ CUB \\
\cline { 3 - 4 } & & 5 -way 1-shot & 5-way 5-shot \\
\hline Fine-tuning (Sun, Lapuschkin, Samek, et al., 2021) & ResNet-10 & $41.98 \pm 0.41$ & $58.75 \pm 0.36$ \\
RelationNet (Sung, Yang, Zhang, et al., 2018) & ResNet-18 & $42.91 \pm 0.78$ & $57.71 \pm 0.73$ \\
LRP-RN (Hu \& Ma, 2022) & ResNet-10 & $42.44 \pm 0.41$ & $59.30 \pm 0.40$ \\
MN+AFA (Chen, Liu, Kira, et al., 2018) & ResNet-10 & $41.02 \pm 0.40$ & $59.46 \pm 0.40$ \\
PDN-PAS (Chen et al., 2023a) & ResNet-18 & $42.41 \pm 0.84$ & $61.25 \pm 0.86$ \\
Baseline++ (Fu, Fu, \& Jiang, 2021) & ResNet-18 & $43.04 \pm 0.60$ & $62.04 \pm 0.76$ \\
Baseline (Fu et al., 2021) & ResNet-18 & - & $65.57 \pm 0.70$ \\
MatchingNet (Vinyals, Blundell, Lillicrap, et al., 2016) & ResNet-18 & $45.59 \pm 0.81$ & $53.07 \pm 0.74$ \\
ProtoNet (Snell et al., 2017) & ResNet-18 & $45.31 \pm 0.78$ & $62.02 \pm 0.70$ \\
GNN (Garcia \& Bruna, 2018) & ResNet-10 & $45.69 \pm 0.68$ & $62.25 \pm 0.65$ \\
GNN+FT (Tseng, Lee, Huang, et al., 2020) & ResNet-10 & $47.47 \pm 0.75$ & $66.98 \pm 0.68$ \\
FDMixup (Gao, Su, Prasad, et al., 2024) & ResNet-10 & $46.38 \pm 0.68$ & $64.71 \pm 0.68$ \\
MIFN (Zhang, Cai, Lin, et al., 2020) & ResNet-12 & $48.21 \pm 0.60$ & $65.33 \pm 0.54$ \\
KLSANet & ResNet-12 & $48.16 \pm 0.64$ & $67.25\pm 0.61$ \\
WATF & ResNet-12 & $\textbf{48.39} \pm \textbf{0.58}$ & $\textbf{68.29}\pm \textbf{0.57}$ \\
\hline
\end{tabular}
\end{table*}

\subsubsection{Cross-domain Few-Shot Classification}

To evaluate the cross-domain generalization of WATF, we conducted experiments in the miniImageNet→CUB setting (see Table \ref{tab2}) and compared it with state-of-the-art methods. The model was trained on 64 base classes from miniImageNet and performance was evaluated on 50 novel classes in the CUB test set. WATF demonstrated significant advantages in this cross-domain scenario, achieving an accuracy of 48.39\% in the 5-way 1-shot setting and 68.92\% in the 5-way 5-shot setting.

It also outperformed classic few-shot methods such as MatchingNet, ProtoNet, RelationNet, and GNN, for example, surpassing ProtoNet by 3.08\% and 6.27\% in 1-shot and 5-shot settings, respectively. Notably, compared to methods tailored for cross-domain scenarios (such as Finetuning, LRP, MN+AFA, baseline, baseline++, GNN+FT, and FDMixup), WATF maintained a lead. For instance, in 1-shot and 5-shot settings, it outperformed the FDMixup method (which advocates using limited labeled target data to guide cross-domain learning) by 2.01\% and 3.58\%, respectively.

\subsection{Ablation Studies}


\begin{table*}[!t]
\caption{The influence of using weighted
adaptive threshold filtering (WATF) strategy.}
\label{tab3}
\centering
\begin{tabular}{cccccccc}
\toprule
\multirow{2}{*}{WATF} & \multirow{2}{*}{Backbone}& \multicolumn{2}{c}{CUB} & \multicolumn{2}{c}{Stanford dogs} & \multicolumn{2}{c}{Stanford cars} \\
\cmidrule{3-4} \cmidrule{5-6} \cmidrule{7-8}
& & 1-shot & 5-shot & 1-shot & 5-shot & 1-shot & 5-shot \\
\midrule
w/o & Conv-4 & $46.84 \pm 0.81$ & $74.92 \pm 0.64$ & $45.41 \pm 0.76$ & $63.51 \pm 0.62$ & $59.84 \pm 0.80$ & $88.65 \pm 0.44$ \\
w & Conv-4 & $65.94 \pm 0.97$ & $79.96 \pm 0.52$ & $56.00 \pm 0.68$ & $73.70 \pm 0.54$ & $57.95 \pm 0.61$ & $81.66 \pm 0.45$ \\
\bottomrule
\end{tabular}
\end{table*}


\subsubsection{Impact of WATFM}

This paper proposes WATF, which innovatively introduces a weighted local descriptor adaptive threshold filtering strategy to improve classification performance. This section investigates the effectiveness of our method in eliminating class-irrelevant noise information by comparing the experimental accuracy using our WATF strategy against that without any processing of local descriptors.

As shown in Table \ref{tab3}, where "w/" and "w/o" denote the use and non-use of the WATF strategy respectively, the results demonstrate that our WATF strategy can eliminate class-irrelevant noise information.
\begin{figure*}[ht]
\centering
\includegraphics[scale=0.6]{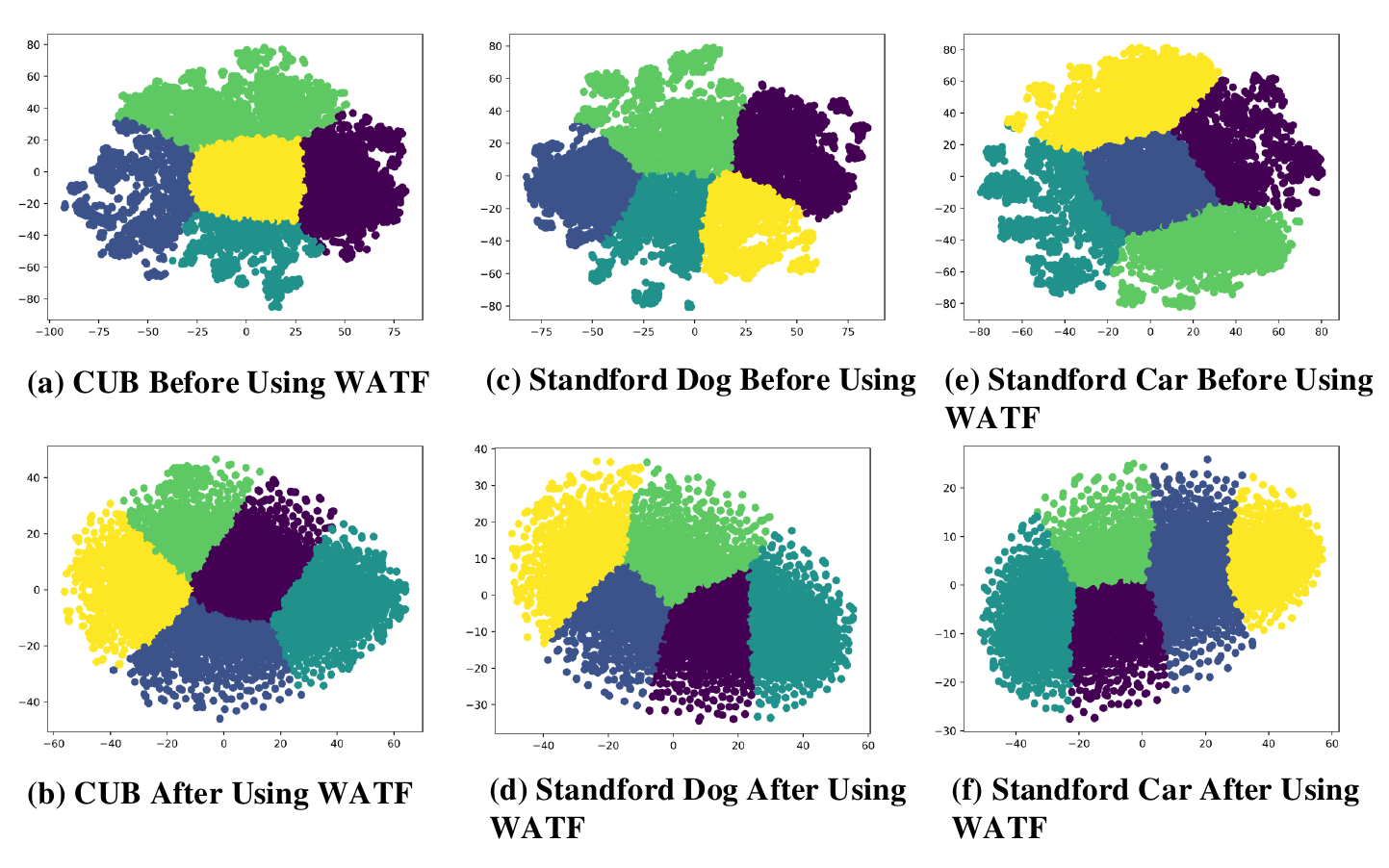}
\caption{Visualization of features before and after applying the WATF operation on three datasets.}
\label{julei}
\end{figure*}
Furthermore, since our WATF module achieves the elimination of noisy local descriptor features, it should produce more effective representations. To better understand the changes in local descriptor feature distribution before and after WATF, we visualize their representations in two-dimensional space using t-SNE technology. Figure \ref{julei} shows the distribution of support set local descriptor features before and after applying WATF on three fine-grained classification datasets. From the visualization, it can be observed that after applying WATF in each class, the local descriptor features become more tightly clustered together, and the class boundaries become clearer. The extensive experimental evaluations in this study confirm that the enhanced inter-class separability contributes to the subsequent $k$-NN classifier's improvement in classification, enhancing feature stability.

\begin{figure*}[ht]
\centering
\includegraphics[scale=0.6]{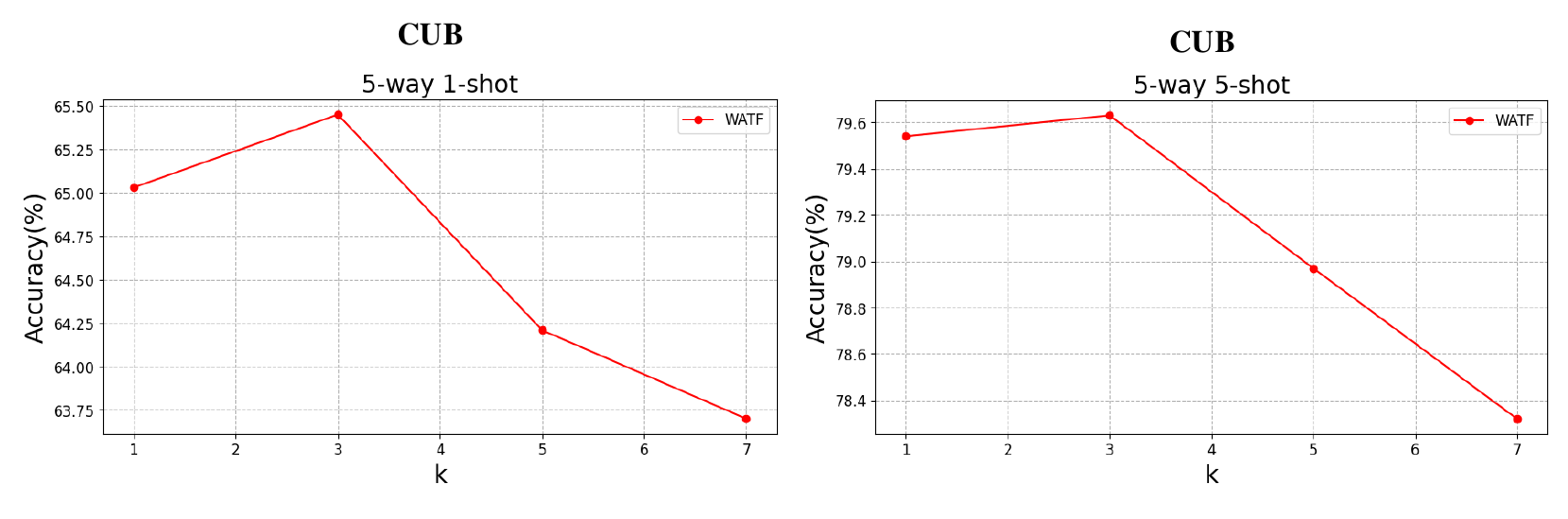}
\caption{ $k$-NN Classifier Parameter Analysis Using The Conv-4 Backbone On The CUB Dataset.}
\label{kvalue}
\end{figure*}

\subsubsection{Impact of Different $k$ Values in $k$-NN Classifier on Experimental Results}

Following the work of DN4 \cite{li2019revisiting}, BDLA \cite{zheng2023bdla}, and DLDA \cite{song2024learning}, we employ the k-nearest neighbors ($k$-NN) model as the classifier to align the similar semantic information between local descriptor features of images. To investigate the impact of different $k$ values on the results of the FAFD-LDWR method, we conducted $k$-NN parameter analysis using the Conv-4 backbone on the CUB dataset. Specifically, we experimented with different k values (i.e., k = 1, 3, 5, 7). The experimental results, as shown in Figure \ref{kvalue}, indicate that, consistent with the conclusions drawn in the DN4 and BDLA studies, the best classification accuracy is achieved when k = 3.

\section{conclusion}
In this study, we propose a effective WATF method to enhance the performance of few-shot learning.

This approach enables the feature extractor to effectively focus on local descriptors relevant to the image class, thereby reducing the interference of class-irrelevant information.

Our weighted adaptive threshold filtering module focuses on class-relevant key information, enhancing image representation and reducing the impact of irrelevant regions. This improves classification accuracy by filtering out irrelevant background descriptors. The method remains simple and lightweight, introducing no additional learnable parameters and maintaining consistency between training and testing phases.

The proposed method is expected to work in other data modalities such as medical images and text data, which will be investigated in future work.

\bibliography{example_paper}

\begin{thebibliography}{10}
\providecommand{\url}[1]{#1}
\csname url@samestyle\endcsname
\providecommand{\newblock}{\relax}
\providecommand{\bibinfo}[2]{#2}
\providecommand{\BIBentrySTDinterwordspacing}{\spaceskip=0pt\relax}
\providecommand{\BIBentryALTinterwordstretchfactor}{4}
\providecommand{\BIBentryALTinterwordspacing}{\spaceskip=\fontdimen2\font plus
\BIBentryALTinterwordstretchfactor\fontdimen3\font minus \fontdimen4\font\relax}
\providecommand{\BIBforeignlanguage}[2]{{%
\expandafter\ifx\csname l@#1\endcsname\relax
\typeout{** WARNING: IEEEtran.bst: No hyphenation pattern has been}%
\typeout{** loaded for the language `#1'. Using the pattern for}%
\typeout{** the default language instead.}%
\else
\language=\csname l@#1\endcsname
\fi
#2}}
\providecommand{\BIBdecl}{\relax}
\BIBdecl

\bibitem{song2023learning}
Q.~Song, S.~Zhou, and L.~Xu, ``Learning more discriminative local descriptors for few-shot learning,'' \emph{arXiv preprint arXiv:2305.08721}, 2023.

\bibitem{li2020more}
H.~Li, L.~Yang, and F.~Gao, ``More attentional local descriptors for few-shot learning,'' in \emph{International Conference on Artificial Neural Networks}.\hskip 1em plus 0.5em minus 0.4em\relax Springer, 2020, pp. 419--430.

\bibitem{zheng2023bdla}
Z.~Zheng, X.~Feng, H.~Yu, X.~Li, and M.~Gao, ``Bdla: Bi-directional local alignment for few-shot learning,'' \emph{Applied Intelligence}, vol.~53, no.~1, pp. 769--785, 2023.

\bibitem{li2019revisiting}
W.~Li, L.~Wang, J.~Xu, J.~Huo, Y.~Gao, and J.~Luo, ``Revisiting local descriptor based image-to-class measure for few-shot learning,'' in \emph{Proceedings of the IEEE/CVF conference on computer vision and pattern recognition}, 2019, pp. 7260--7268.

\bibitem{sun2024klsanet}
Z.~Sun, W.~Zheng, and P.~Guo, ``Klsanet: Key local semantic alignment network for few-shot image classification,'' \emph{Neural Networks}, p. 106456, 2024.

\bibitem{zhou2024global}
J.~Zhou and Q.~Cai, ``Global and local representation collaborative learning for few-shot learning,'' \emph{Journal of Intelligent Manufacturing}, vol.~35, no.~2, pp. 647--664, 2024.

\bibitem{hao2021global}
F.~Hao, F.~He, J.~Cheng, and D.~Tao, ``Global-local interplay in semantic alignment for few-shot learning,'' \emph{IEEE Transactions on Circuits and Systems for Video Technology}, vol.~32, no.~7, pp. 4351--4363, 2021.

\bibitem{finn2017model}
C.~Finn, P.~Abbeel, and S.~Levine, ``Model-agnostic meta-learning for fast adaptation of deep networks,'' in \emph{International conference on machine learning}.\hskip 1em plus 0.5em minus 0.4em\relax PMLR, 2017, pp. 1126--1135.

\bibitem{lee2019meta}
K.~Lee, S.~Maji, A.~Ravichandran, and S.~Soatto, ``Meta-learning with differentiable convex optimization,'' in \emph{Proceedings of the IEEE/CVF conference on computer vision and pattern recognition}, 2019, pp. 10\,657--10\,665.

\bibitem{wang2024fast}
M.~Wang, Q.~Gong, Q.~Wan, Z.~Leng, Y.~Xu, B.~Yan, H.~Zhang, H.~Huang, and S.~Sun, ``A fast interpretable adaptive meta-learning enhanced deep learning framework for diagnosis of diabetic retinopathy,'' \emph{Expert Systems with Applications}, vol. 244, p. 123074, 2024.

\bibitem{snell2017prototypical}
J.~Snell, K.~Swersky, and R.~Zemel, ``Prototypical networks for few-shot learning,'' \emph{Advances in neural information processing systems}, vol.~30, 2017.

\bibitem{vinyals2016matching}
O.~Vinyals, C.~Blundell, T.~Lillicrap, D.~Wierstra \emph{et~al.}, ``Matching networks for one shot learning,'' \emph{Advances in neural information processing systems}, vol.~29, 2016.

\bibitem{huang2021local}
H.~Huang, Z.~Wu, W.~Li, J.~Huo, and Y.~Gao, ``Local descriptor-based multi-prototype network for few-shot learning,'' \emph{Pattern Recognition}, vol. 116, p. 107935, 2021.

\bibitem{qi2022task}
Y.~Qi, H.~Sun, N.~Liu, and H.~Zhou, ``A task-aware dual similarity network for fine-grained few-shot learning,'' in \emph{Pacific Rim International Conference on Artificial Intelligence}.\hskip 1em plus 0.5em minus 0.4em\relax Springer, 2022, pp. 606--618.

\bibitem{sung2018learning}
F.~Sung, Y.~Yang, L.~Zhang, T.~Xiang, P.~H. Torr, and T.~M. Hospedales, ``Learning to compare: Relation network for few-shot learning,'' in \emph{Proceedings of the IEEE conference on computer vision and pattern recognition}, 2018, pp. 1199--1208.

\bibitem{chen2024featwalk}
D.~Chen, J.~Zhang, W.-S. Zheng, and R.~Wang, ``Featwalk: Enhancing few-shot classification through local view leveraging,'' in \emph{Proceedings of the AAAI Conference on Artificial Intelligence}, vol.~38, no.~2, 2024, pp. 1019--1027.

\bibitem{fu2021meta}
Y.~Fu, Y.~Fu, and Y.-G. Jiang, ``Meta-fdmixup: Cross-domain few-shot learning guided by labeled target data,'' in \emph{Proceedings of the 29th ACM international conference on multimedia}, 2021, pp. 5326--5334.

\bibitem{hu2022adversarial}
Y.~Hu and A.~J. Ma, ``Adversarial feature augmentation for cross-domain few-shot classification,'' in \emph{European conference on computer vision}.\hskip 1em plus 0.5em minus 0.4em\relax Springer, 2022, pp. 20--37.

\bibitem{gao2024few}
R.~Gao, H.~Su, S.~Prasad, and P.~Tang, ``Few-shot classification with multisemantic information fusion network,'' \emph{Image and Vision Computing}, vol. 141, p. 104869, 2024.

\bibitem{tseng2020cross}
H.-Y. Tseng, H.-Y. Lee, J.-B. Huang, and M.-H. Yang, ``Cross-domain few-shot classification via learned feature-wise transformation,'' \emph{arXiv preprint arXiv:2001.08735}, 2020.

\bibitem{chen2021meta}
Y.~Chen, Z.~Liu, H.~Xu, T.~Darrell, and X.~Wang, ``Meta-baseline: Exploring simple meta-learning for few-shot learning,'' in \emph{Proceedings of the IEEE/CVF international conference on computer vision}, 2021, pp. 9062--9071.

\bibitem{sun2021explanation}
J.~Sun, S.~Lapuschkin, W.~Samek, Y.~Zhao, N.-M. Cheung, and A.~Binder, ``Explanation-guided training for cross-domain few-shot classification,'' in \emph{2020 25th international conference on pattern recognition (ICPR)}.\hskip 1em plus 0.5em minus 0.4em\relax IEEE, 2021, pp. 7609--7616.

\bibitem{dong2021learning}
C.~Dong, W.~Li, J.~Huo, Z.~Gu, and Y.~Gao, ``Learning task-aware local representations for few-shot learning,'' in \emph{Proceedings of the twenty-ninth international conference on international joint conferences on artificial intelligence}, 2021, pp. 716--722.

\bibitem{rusu2018meta}
A.~A. Rusu, D.~Rao, J.~Sygnowski, O.~Vinyals, R.~Pascanu, S.~Osindero, and R.~Hadsell, ``Meta-learning with latent embedding optimization,'' \emph{arXiv preprint arXiv:1807.05960}, 2018.

\bibitem{triantafillou2019meta}
E.~Triantafillou, T.~Zhu, V.~Dumoulin, P.~Lamblin, U.~Evci, K.~Xu, R.~Goroshin, C.~Gelada, K.~Swersky, P.-A. Manzagol \emph{et~al.}, ``Meta-dataset: A dataset of datasets for learning to learn from few examples,'' \emph{arXiv preprint arXiv:1903.03096}, 2019.

\bibitem{wang2022global}
H.~Wang, Y.~Wang, R.~Sun, and B.~Li, ``Global convergence of maml and theory-inspired neural architecture search for few-shot learning,'' in \emph{Proceedings of the IEEE/CVF conference on computer vision and pattern recognition}, 2022, pp. 9797--9808.

\bibitem{baik2021meta}
S.~Baik, J.~Choi, H.~Kim, D.~Cho, J.~Min, and K.~M. Lee, ``Meta-learning with task-adaptive loss function for few-shot learning,'' in \emph{Proceedings of the IEEE/CVF international conference on computer vision}, 2021, pp. 9465--9474.

\bibitem{oreshkin2018tadam}
B.~Oreshkin, P.~Rodr{\'\i}guez~L{\'o}pez, and A.~Lacoste, ``Tadam: Task dependent adaptive metric for improved few-shot learning,'' \emph{Advances in neural information processing systems}, vol.~31, 2018.

\bibitem{li2019distribution}
W.~Li, J.~Xu, J.~Huo, L.~Wang, Y.~Gao, and J.~Luo, ``Distribution consistency based covariance metric networks for few-shot learning,'' in \emph{Proceedings of the AAAI conference on artificial intelligence}, vol.~33, no.~01, 2019, pp. 8642--8649.

\bibitem{satorras2018few}
V.~G. Satorras and J.~B. Estrach, ``Few-shot learning with graph neural networks,'' in \emph{International conference on learning representations}, 2018.

\bibitem{abdelaziz2023learn}
M.~Abdelaziz and Z.~Zhang, ``Learn to aggregate global and local representations for few-shot learning,'' \emph{Multimedia Tools and Applications}, vol.~82, no.~21, pp. 32\,991--33\,014, 2023.

\bibitem{song2024learning}
Song \emph{et~al.}, ``Learning more discriminative local descriptors with parameter-free weighted attention for few-shot learning,'' vol.~35, no.~4, p.~71, 2024.

\end{thebibliography}
\bibliographystyle{IEEEtran}

\vfill

\end{document}